\documentclass[lettersize,journal]{IEEEtran}
\usepackage{amsmath}
\usepackage{amsfonts}
\usepackage{algorithmic}
\usepackage{algorithm}
\usepackage{array}
\usepackage[caption=false,font=normalsize,labelfont=sf,textfont=sf]{subfig}
\usepackage{textcomp}
\usepackage{stfloats}
\usepackage{url}
\usepackage{verbatim}
\usepackage{graphicx}
\usepackage{multirow}
\usepackage{multicol}
\usepackage{cite}
\usepackage{orcidlink}

\begin{document}

\title{SALT: A Flexible Semi-Automatic Labeling Tool for General LiDAR Point Clouds with Cross-Scene Adaptability and 4D Consistency}

\author{Yanbo Wang \orcidlink{0000-0002-9098-5071}, Yongtao Chen \orcidlink{0009-0000-6910-5503}, Chuan Cao \orcidlink{0009-0005-4857-3806}, Tianchen Deng \orcidlink{0000-0002-4368-7936},  Wentao Zhao \orcidlink{0000-0002-4125-239X}, Jingchuan Wang \orcidlink{0000-0002-1943-1535}, ~\IEEEmembership{Senior Member,~IEEE,} Weidong Chen \orcidlink{0000-0001-8757-0679}, ~\IEEEmembership{Member,~IEEE,}
        
\thanks{Yanbo Wang, Yongtao Chen, Chuan Cao, Tianchen Deng, Wentao Zhao, Jingchuan Wang,
and Weidong Chen are with the Institute of Medical Robotics, Department of
Automation, Shanghai Jiao Tong University, Shanghai 200240, China. Yanbo Wang and Yongtao Chen contributed equally to this work. Jingchuan Wang (jchwang@sjtu.edu.cn) and Weidong Chen (wdchen@sjtu.edu.cn) are the corresponding authors.}}

% \thanks{This work is supported by the National Key R\&D Program of China (Grant 2020YFC2007500), and the Science and Technology Commission of Shanghai Municipality (Grant 20DZ2220400). }}% <-this % stops a space
%\thanks{Manuscript received April 19, 2021; revised August 16, 2021.}}

\markboth{Journal of \LaTeX\ Class Files,~Vol.~14, No.~8, August~2021}%
{Shell \MakeLowercase{\textit{et al.}}: A Sample Article Using IEEEtran.cls for IEEE Journals}


\maketitle

\begin{abstract}
We propose a flexible Semi-Automatic Labeling Tool (SALT) for general LiDAR point clouds with cross-scene adaptability and 4D consistency. Unlike recent approaches that rely on camera distillation, SALT operates directly on raw LiDAR data, automatically generating pre-segmentation results. To achieve this, we propose a novel zero-shot learning paradigm, termed data alignment, which transforms LiDAR data into pseudo-images by aligning with the training distribution of vision foundation models. Additionally, we design a 4D-consistent prompting strategy and 4D non-maximum suppression module to enhance SAM2, ensuring high-quality, temporally consistent presegmentation. SALT surpasses the latest zero-shot methods by 18.4\% PQ on SemanticKITTI and achieves nearly 40\~{}50\% of human annotator performance on our newly collected low-resolution LiDAR data and on combined data from three LiDAR types, significantly boosting annotation efficiency. We anticipate that SALT's open-sourcing will catalyze substantial expansion of current LiDAR datasets and lay the groundwork for the future development of LiDAR foundation models. Code is available at \url{https://github.com/Cavendish518/SALT}.
\end{abstract}

\begin{IEEEkeywords}
Zero-shot learning, annotation tool, LiDAR panoptic segmentation, 4D panoptic LiDAR segmentation, data alignment.
\end{IEEEkeywords}

\section{Introduction}
\label{sec:intro}
As increasingly large and high-quality datasets become available, the artificial intelligence and robotics communities are experiencing unprecedented growth. The assembly of large-scale text corpora has driven advancements in large language models (LLM)~\cite{chowdhery2023palm,touvron2023llama,achiam2023gpt,team2023gemini}, while extensive image datasets have accelerated the development of vision foundation models (VFM)~\cite{kirillov2023segment,ravi2024sam,zou2023generalized,zou2024segment,radford2021learning}. However, the foundational models for 3D LiDAR, a critical modality for autonomous driving and robotics \cite{sirohi2021efficientlps}, lag behind in development due to the scarcity of current datasets~\cite{hu2021towards,unal2022scribble,caesar2020nuscenes,xiao2021pandaset,behley2019semantickitti,wang2024sfpnet,sun2020scalability,jiang2021rellis,pan2020semanticposs,yan2024benchmarking}.

\begin{figure*}[t]
  \centering
\includegraphics[width=1.0\linewidth]{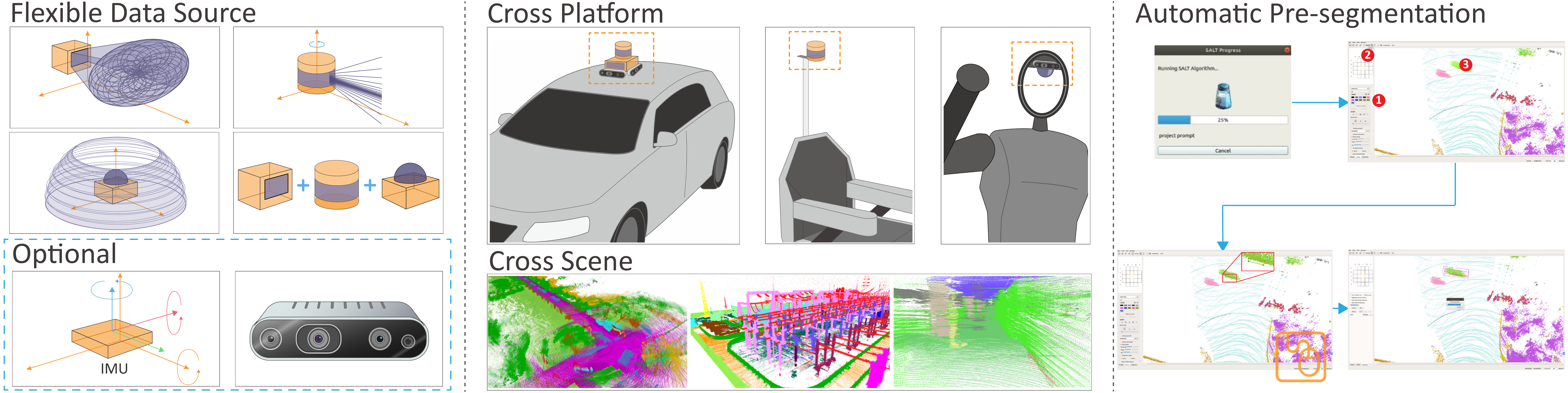}
        \caption{Overview of SALT: Flexible data sources, cross-platform adaptability, and automatic presegmentation workflow.}
        \label{fig1:overall}
\end{figure*}

This challenge largely stems from the high cost of qualified annotation \cite{10891936}. For instance, annotating \textbf{40,000} frames can require up to \textbf{1,700} hours of labor~\cite{behley2019semantickitti}. Therefore, despite the widespread availability of various types of LiDAR sensors, the scale of public datasets remains constrained by cost and quality considerations. A promising approach to alleviate the annotation burden is to utilize a presegmentation model. However, an effective presegmentation model relies on large-scale dataset training, which seems to put us in a chicken-and-egg dilemma.

One potential solution is to implement an interactive segmentation model~\cite{kontogianni2023interactive, sun2023click, han2024scale}, incorporating human prior information to aid in segmentation. Another approach involves knowledge distillation~\cite{liu2024segment,ovsep2025better,peng2025learning} from VFM. However, the former method relies on manual clicks and does not completely reduce human workload. The latter method heavily depends on calibrated cameras. More importantly, both methods rely on training of a certain small-scale LiDAR dataset, limiting their zero-shot capabilities.  As a result, presegmentation performance suffers on newly collected data (especially for different types of LiDAR), thereby increasing the manual cost of annotation.

To overcome these limitations, we propose a new paradigm called data alignment. Inspired by decision boundary studies in adversarial attacks, data alignment turns the inherent vulnerability of neural networks into an advantage. By transforming LiDAR data into pseudo-images aligned with the VFM training dataset, we satisfy decision boundary conditions, enabling accurate cross-domain segmentation. We train a deep clustering network on the VFM dataset to measure the distance between pseudo-images and the dataset, optimizing modality transformation by retaining point cloud information and minimizing this distance. To enhance robustness in VFM, we propose a pseudo-color mechanism. Additionally, a 4D-consistent prompting strategy and 4D non-maximum suppression (NMS) ensure high-quality, temporally consistent panoptic presegmentation. Combining these elements, we develop a flexible Semi-Automatic Labeling Tool (SALT) for general LiDAR point clouds with cross-scene adaptability and 4D consistency.

We evaluate the automatic segmentation performance of our tool on four cross-scene public LiDAR segmentation benchmarks (nuScenes~\cite{caesar2020nuscenes,fong2022panoptic}, SemanticKITTI ~\cite{behley2019semantickitti,behley2021benchmark}, SemanticKITTI-16~\cite{yan2024benchmarking}, and S.MID~\cite{wang2024sfpnet}) with different LiDAR sensors. Our tool achieves an improvement of 18.4\% $PQ$ (SemanticKITTI) and 3.0\% $PQ$ (nuScenes) over state-of-the-art (SOTA) zero-shot methods. Moreover, SALT achieves 31.5\% $LSTQ$ on SemanticKITTI, 46.1\% $mIoU$ on S.MID and 28.2\% $mIoU$ on SemanticKITTI-16. We also conduct tests on two custom-built platforms: one designed for annotating data from a low-cost, 16-beam LiDAR in indoor environments, and another configured to handle data from a combination of three LiDAR types in outdoor environments. Our tool achieve nearly 40\~{}50\% of human annotator performance while boosting annotation efficiency by approximately 6 times.

We summarize our contributions as below:

\begin{itemize}

\item[$\bullet$] We propose a self-adaptive and zero-shot framework for segmenting any LiDAR. To achieve this, a novel data alignment paradigm for cross-domain knowledge transfer is proposed and a 4D-consistent prompting strategy is formulated.
\item[$\bullet$] Our method demonstrates robust zero-shot capabilities and high adaptability across various sensor combinations, diverse scenes, and platform movement speeds in data collection, which is supported by SOTA zero-shot presegmentation results achieved on four public benchmarks and two self-built platforms.
\item[$\bullet$] We develop the first training-free semi-automatic labeling tool focused on general LiDAR data, which operates flexibly with or without camera and IMU input, and supports any type of LiDAR (mechanical spinning, solid-state, hybrid-solid) or combinations of them.

\end{itemize}

\section{Related Work}
\subsection{LiDAR Point Cloud Segmentation}
Given its importance in robotics and autonomous driving applications, 3D LiDAR segmentation has experienced flourishing development. Although segmentation tasks can be categorized into semantic level, instance level, panoptic level and 4D panoptic level, they can generally be divided into four types based on the input to the networks. Point-based~\cite{qi2017pointnet++,hu2020randla,thomas2019kpconv,qian2022pointnext,zhao2021point,zhang2023pids,aygun20214d,gasperini2021panoster,marcuzzi2023mask4d}, projection-based~\cite{xu2020squeezesegv3,zhang2020polarnet,puy2023using,kong2023rethinking,ando2023rangevit,xu2023frnet,cortinhal2020salsanext,zhou2021panoptic,chen2021polarstream,sirohi2021efficientlps}, voxel-based~\cite{choy20194d,graham20183d,zhu2021cylindrical,liu2022less,lai2023spherical,li2023less,li2025rapid,hong2021lidar} and multi-modality-based~\cite{xu2021rpvnet,yan20222dpass,liu2023uniseg,cheng20212,zhuang2021perception,hong2021lidar}. Despite the notable success of LiDAR point cloud segmentation, the effectiveness of automatic annotation using them directly in a zero-shot manner remains unsatisfactory.

Recently, PPT~\cite{wu2024towards} and PTv3~\cite{wu2024point} advance multi-dataset synergistic training through an effective pretraining approach called Point Prompt Training. SFPNet~\cite{wang2024sfpnet} expands dataset variety from different types of LiDAR data and introduces sparse focal point modulation to handle these variations. COLA~\cite{10891936} performs multi-source domain generalization by introducing the same coarse label to multiple datasets. This pretraining method enables the network to overcome the domain differences between different types of LiDAR to a certain extent and makes the subsequent fintuning of a single dataset perform better. While these recent works have laid foundational steps toward scaling up LiDAR-based segmentation, they still rely on training within specific datasets, lacking a truly universal approach similar to SAM~\cite{kirillov2023segment}. Further progress is hindered by the small scale of existing datasets and complicated dataset consolidation due to disparities in point cloud distribution across different LiDAR types. These limitations restrict the potential for scaling up, leaving the field waiting for a breakthrough moment.

\subsection{Rethinking Feature Alignment via VFM}
\label{sec:rw}
\begin{figure}[t]
    \centering
    \includegraphics[width=1 \linewidth]{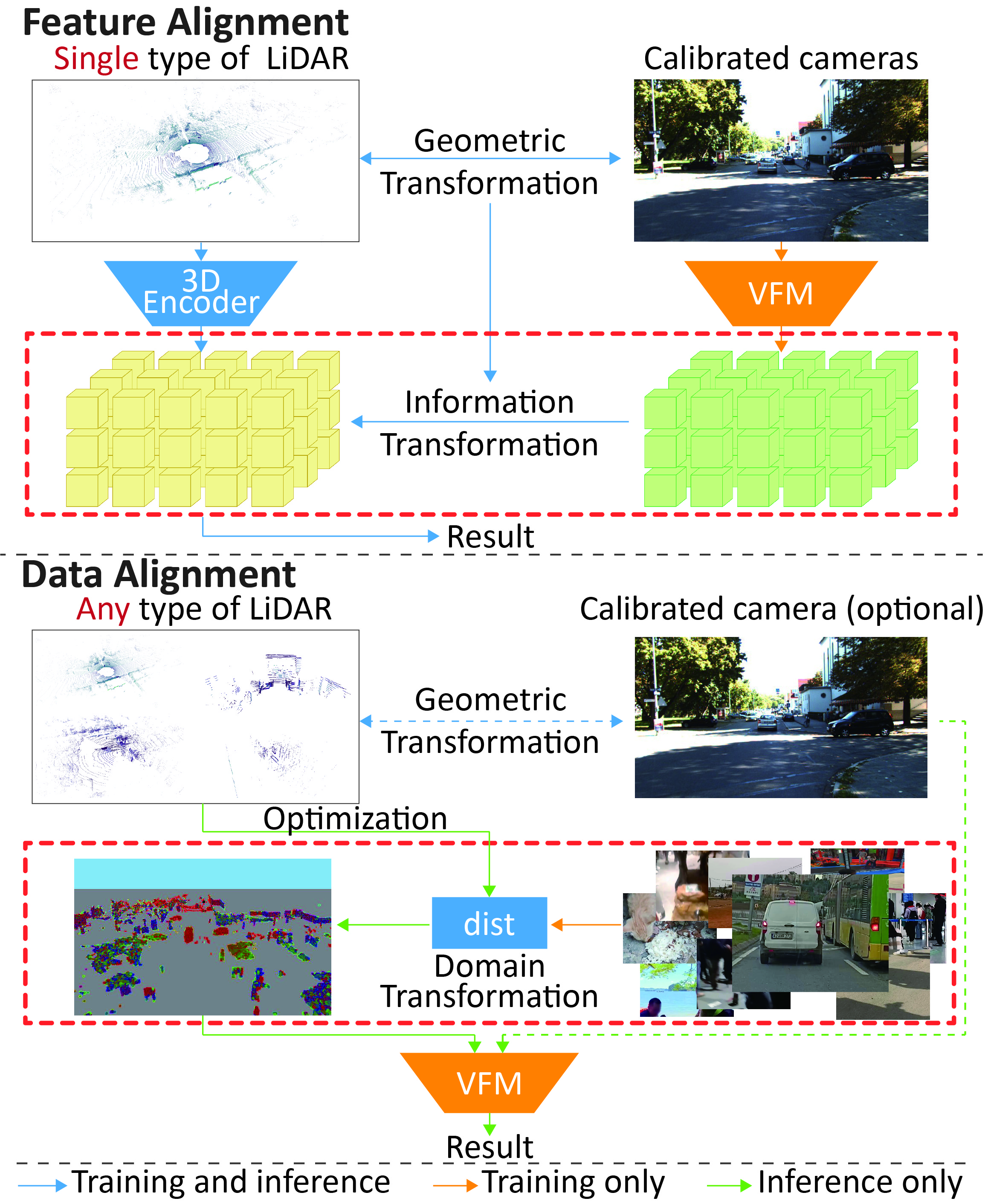}
    \caption{Feature Alignment vs Data Alignment. Both paradigms utilize VFM knowledge, but they differ in the transfer process. Feature alignment transfers knowledge after encoding the source domain, meaning the segmentation performance of the 3D encoder is constrained by the source domain size during training. In contrast, data alignment directly transforms the source domain into the target domain with respect to decision boundary conditions, fully preserving VFM knowledge to achieve true zero-shot capability. This is crucial for annotation tools that need to adapt across diverse data collection platforms.}
    \label{fig2:alignment}
\end{figure}

In order to overcome the limited scalability problem in the LiDAR segmentation field, cutting-edge works~\cite{liu2024segment,zhang2023sam3d,guo2024sam2point,ovsep2025better,peng2025learning} introduce VFM to expand knowledge domain. Seal~\cite{liu2024segment} introduces VFM-assisted contrastive learning for pretraining. Peng et al.~\cite{peng2025learning} utilize extensive knowledge from SAM~\cite{kirillov2023segment} to unify feature representations across various 3D domains. SAL~\cite{ovsep2025better} employs SAM~\cite{kirillov2023segment} and MaskCLIP~\cite{ding2023open} to build a pseudo-labeling engine that facilitates model training without manual supervision. As illustrated in Fig.~\ref{fig2:alignment}, these methods rely on calibrated cameras from each dataset to perform feature alignment through a distillation process. During the development of a general labeling tool, we identify several issues. First and foremost, these methods become ineffective when camera observations are insufficient or absent. This is evident from the results and analyses in studies~\cite{ovsep2025better,peng2025learning}, which demonstrate performance differences between nuScenes~\cite{caesar2020nuscenes} and  SemanticKITTI~\cite{behley2019semantickitti} driven by varying camera coverage. Second, their (pre)training process is limited to a single type of LiDAR, which hinders their capability~\cite{wang2024sfpnet} as true zero-shot labeling models like those in the image domain. Additionally, inherent issues in feature alignment methods, such as data quantity imbalance and information loss during projection, constrain their performance in the labeling process. These problems motivate us to introduce data alignment to unleash the potential of SAM~\cite{kirillov2023segment,ravi2024sam} and build a training-free general LiDAR labeler that supports absence of camera modality.

\subsection{Labeler and Interactive Segmentation}
LiDAR annotation tools can generally be categorized into two types: single-frame annotation and accumulated-frame annotation. SUSTech~\cite{sustech} is a typical example of single-frame annotation, while LABELER~\cite{behley2019semantickitti} represents accumulated-frame annotation. Given that accumulated-frame annotation is compatible with single-frame annotation, we develop an automatic segmentation function based on LABELER. Additionally, another type of research focuses on interactive segmentation~\cite{yue2023agile3d, kontogianni2023interactive, sun2023click, han2024scale, interactive4d}, which achieves segmentation results through multiple clicks on one or more objects. However, these approaches are limited by the scale of training dataset and cannot handle zero-shot presegmentation tasks on different types of LiDAR. Moreover, we believe that obtaining presegmentation results through click-based annotation for each frame or every few frames is highly inefficient. To reduce the workload of the annotators, we designed an effective fully automatic prompting method. Annotators only need to fine-tune segmentation results in 3D space and assign semantic or refine instance labels, significantly reducing the workload and paving the way for expanding the LiDAR dataset.

\section{Method}
\label{sec:method}
\subsection{Problem Statement and System Overview}
Given a sequence of data, the input of our system for each frame $t$ is the unlabeled and unordered LiDAR data $L_{t}^{k} \in \mathbb{R}^{N \times 4} $ from each LiDAR $k$, along with \textbf{optional} calibrated camera data $C_{t}^{o}  \in \mathbb{R}^{H \times W \times 3}$ from each camera $o$ and IMU data $M_{t} \in \mathbb{R}^{6}$. Our goal is to provide multi-frame consistent presegmentation labels $Y_{t}^{k} \in \mathbb{R}^{N \times 1}$ within an integrated user interface. This enables users, regardless of expertise, to directly handle raw data from any platform. After automatically obtaining satisfactory presegmentation results, only minimal effort is required to refine them manually by adding semantic or instance labels, fine-tuning boundaries, and merging segments as necessary.

\begin{figure*}[t]
    \centering
    \includegraphics[width=1.0\linewidth]{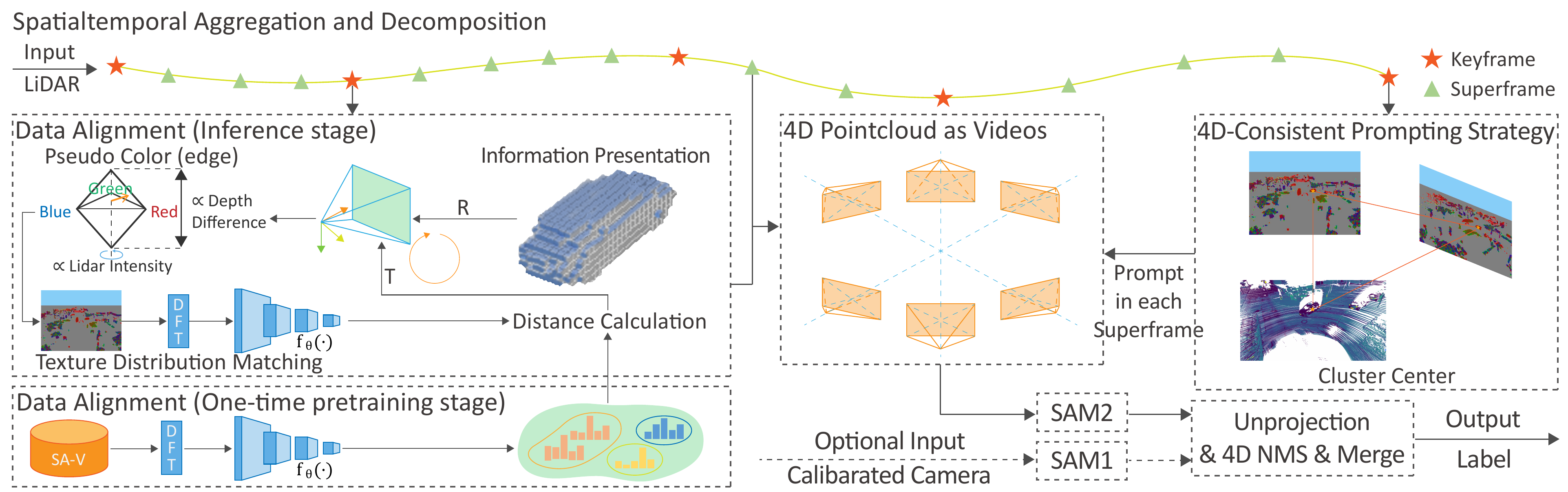}
    \caption{Pipeline of SALT. (1) Spatiotemporal aggregation and decomposition form the foundation of the entire tool, especially supporting core components: \textbf{\textit{Keyframe}} and \textbf{\textit{Superframe}}.  (2) Pseudo-camera is essential for \textbf{data alignment}, effectively aligning the \textbf{texture distribution} of the SAM2 domain while maximizing 4D point cloud \textbf{information}. Pseudo-color provides consistent color information that supports \textbf{edge enhancement} and \textbf{object consistency} within sequence. (3) \textbf{4D-consistent prompting strategy} and \textbf{4D NMS} effectively integrates segmentation results from different perspectives, frames.}
    \label{fig3:pipline}
\end{figure*}

The pipeline of our system is shown in Fig.~\ref{fig3:pipline}. We adopt a hierarchical approach. First, we perform spatiotemporal aggregation and then decompose objects and ground into two groups (Sec.~\ref{sec3.2}). Then, we build a data alignment strategy to convert the modality (Sec.~\ref{sec3.3}). Finally, we use the 4D-consistent prompting strategy with SAM2 \cite{ravi2024sam} to obtain the presegmentation result and convert it back to point cloud for users to finetune (Sec.~\ref{sec3.4}).

\subsection{Spatiotemporal Aggregation and Decomposition}
\label{sec3.2}
Considering the strong spatiotemporal properties of 3D LiDAR data sequence, we first integrate a robust SLAM system~\cite{xu2022fast,lin2022r} to estimate inter-frame poses, accumulate point clouds, and designate the \textit{key} stamps based on pose changes, which will be used in Sec.~\ref{sec3.3} and Sec.~\ref{sec3.4}. While mainstream interactive segmentation and distilled VFM methods typically process single frames, we propose that, from a general labeling perspective, point cloud accumulation effectively mitigates point sparsity, reduces variations across different types of LiDAR, enhances inter-frame consistency, and improves both efficiency and labeling accuracy. We then modify Patchwork++~\cite{patchwork++} to fit the ground (or ceilings), preparing two distinct point cloud sets for segmentation:  \textbf{$L_{object}$}, and \textbf{$L_{ground}$}. Our algorithm design primarily focuses on \textbf{$L_{object}$}. For each frame in $L_{object}$, we construct an accumulated frame called \textit{Superframe}, and voxelize it to obtain $V_{object}$. \textit{Superframe} at \textit{key} stamps is called \textit{Keyframe}. Note that we classify trees and other similar categories into set \textbf{$L_{object}$}, which are typically categorized as \textit{stuff} in traditional panoptic segmentation. This operation will be addressed in Sec.~\ref{metric}.

\subsection{Data Alignment}
\label{sec3.3}
Directly projecting the LiDAR point cloud onto a 2D plane, e.g., image plane or birds eye view (BEV)~\cite{zhang2023sam3d,guo2024sam2point} yields poor performance with VFM like SAM/SAM2, even when denser projections are achieved by accumulating point clouds. This limitation~\cite{wu2023medical,cen2023sad} is fundamentally due to the mismatch between the target data and the training data distribution of SAM/SAM2. Therefore, we introduce the data alignment paradigm as illustrated in Fig.~\ref{fig2:alignment} to fix this problem. This raises the question of what factors most significantly impact the accuracy of VFM. Recent studies~\cite{wang2024empirical} highlight \textbf{texture} and \textbf{edge} information as primary contributors, with color providing some additional influence. Therefore, our data alignment process will focus on these key aspects to make the decision boundary of SAM2 work. Algorithm~\ref{alg:algorithm1} outlines the pseudo-code for our data alignment process, which we will further elaborate on in the following sections.

\subsubsection{Self-Adaptive Domain Transformation}
\label{domain}
The first goal is to ensure the texture of the pseudo-image closely matches the distribution of SAM2's training set (SA-V), thereby preserving SAM2's performance. This is achieved by developing a deep clustering network to classify images in the SA-V dataset and minimizing the distance between the pseudo-image and the dataset during subsequent optimization. The second goal is to retain as much point cloud information as possible to ensure the usability of the segmentation results, which is accomplished by maximizing the number of projection points on the pseudo-image plane.

\noindent \textbf{Metric Network through Deep Fast Clustering.} We randomly sample frames from each video in the SA-V dataset. Then, we construct a smaller representative dataset $D_{sample}$, by cropping and selecting images with the preset size. Since the frequency domain is the optimal representation for texture and edge information, we apply Fourier transform to obtain a representative set of frequency-domain features:

\begin{align}
\begin{split}
   F &= \{F_{i} = \xi(D_i) = \tfrac{|DFT(D_{i})|}{\max(|DFT(D_{i})|)},  D_{i} \in D_{sample}\},
\end{split}
\label{eq:freq}
\end{align}

where $| \cdot |$ calculates the magnitude of the frequency domain sample.

\begin{figure}[t]
    \centering
    \includegraphics[width=1 \linewidth]{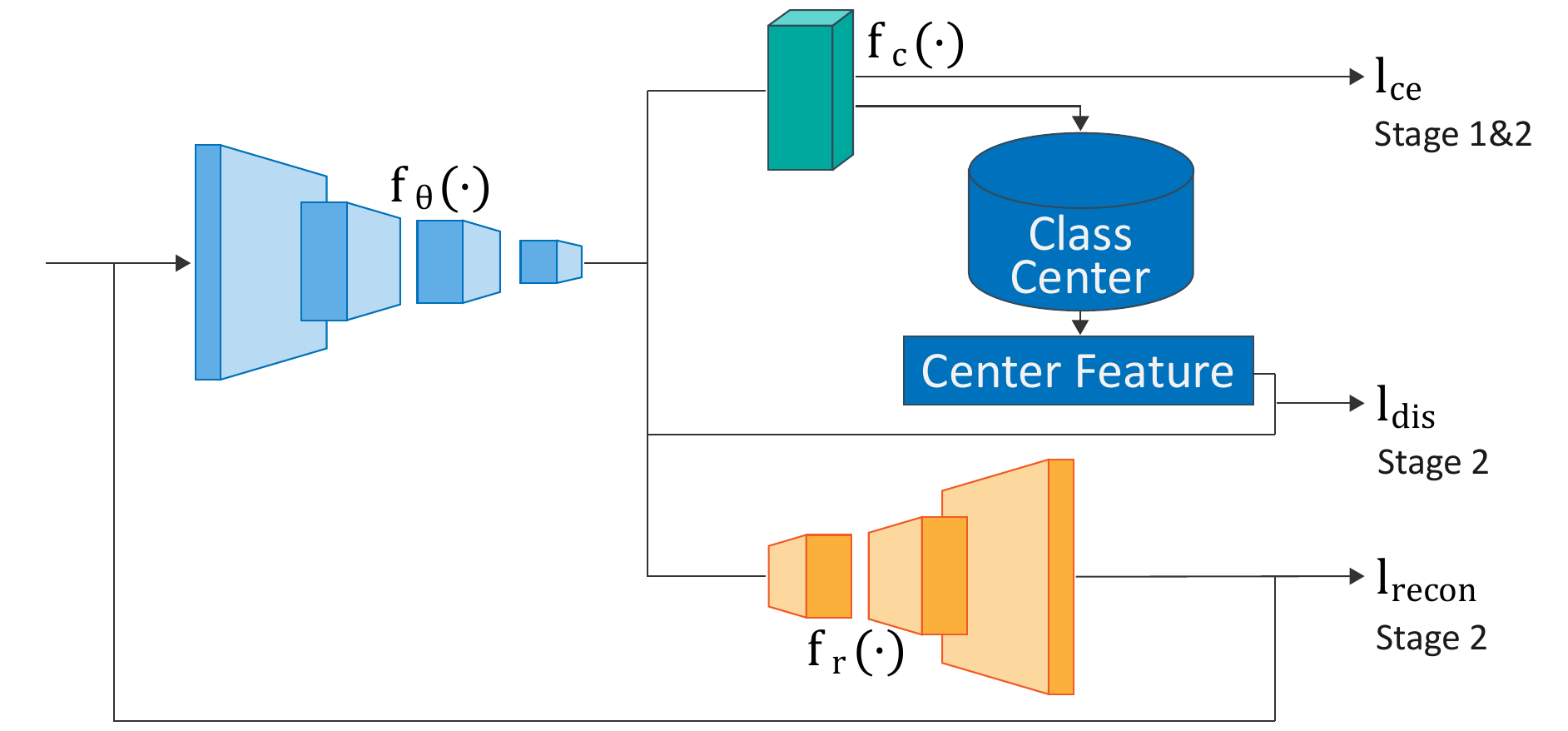}
    \caption{Illustration of our deep fast clustering network. We simply employ ResNet~\cite{he2016deep} as backbone for $f_{\theta}(\cdot)$ and $f_{r}(\cdot)$ and MLP for $f_{c}(\cdot)$. Class center feature are recalculated after each training epoch.}
    \label{fig:supclus}
\end{figure}

Given the representation set $F$ as input, we want to find a function $f_{\theta}( \cdot)$ to project frequency-domain information for the purpose of measuring distances between images. We build $f_{\theta}(\cdot)$ as a metric network for latter optimization problem. Unlike conventional self-supervised frameworks~\cite{caron2018deep}, our deep clustering network employs a two-stage training pipeline to accelerate the training process. Fig. \ref{fig:supclus} illustrates our deep fast clustering network.

In the first stage, we perform pretraining, where the pseudo-labels are generated by histogram statistics and $k$-means \cite{hartigan1979k}:

\begin{align}
    \gamma _{i,k} &=  mean(\mathbb{I}_{[m_k, m_{k+1})} (Freq (F_i)) \odot F_i) \label{eq: HK}, \\
\Gamma _{i} &= [\gamma_{i,k=0},\gamma_{i,k=1},...,\gamma_{i,k=K}] \label{eq:HI} ,
\end{align}
where $[m_k, m_{k+1})$ defines the $k$-${th}$ frequency magnitude bin. $\odot$ is element wise product. $\mathbb{I}( \cdot )$ is the indicator function, which equals 1 if the frequency of $F_i$ falls in the interval and 0 otherwise.  Using $L2$ distance and $k$-means, we cluster the dataset based on $\Gamma _{i}$.

During the first stage, the network is trained via cross-entropy loss $l_{ce}$ with fixed pseudo-labels. In the second stage, the pseudo-labels are iteratively updated, but only at the end of each training epoch. The pseudo-labels are reassigned by performing $k$-means on the features projected from $f_{\theta}( \cdot)$.

\begin{equation}
    l_{ce} = -\frac{1}{N} \sum_{n=1}^{N}\sum_{cls=1}^{Cls}LS(f_c(f_{\theta}(F_{n,cls})))y_{n,cls},  \label{eq:lossce}
\end{equation}
where $f_c(\cdot)$ is the segmentation head. $LS(\cdot)$ represents the log soft-max operation and $y_{n,cls}$ is the one-hot form for pseudo-labels.

In order to reduce the inter-class variance, we also designed discrimination loss $l_{dis}$. 

\begin{equation}
    l_{dis} = \frac{1}{N} \sum_{n=1}^{N} ||f_{\theta}(F_{n,cls=k}) - fea_{cls=k}||_{2}^2,  \label{eq:dis}
\end{equation}
where $fea_{cls=k}$ represents the feature center of the corresponding cluster $k$ calculated at previous training epoch.

We also add reconstruction head and calculate the mean square error reconstruction loss $l_{recon}$. 

\begin{equation}
    l_{recon} = \frac{1}{N} \sum_{n=1}^{N} ||f_r(f_{\theta}(F_{n})) - \xi(x_{n})||_{2}^2,  \label{eq:recon}
\end{equation}
where $f_r(\cdot)$ is the reconstruction head.

The function $f_{\theta}( \cdot)$ trained using Eq.~\eqref{eqloss1} and Eq.~\eqref{eqloss2} can help us determine whether the pseudo-image can be correctly segmented by the decision boundary.

\begin{align}
    loss_{stage1} & = l_{ce} \label{eqloss1}, \\
    loss_{stage2} & = {\omega}_{1}l_{ce} + {\omega}_{2}l_{recon} + {\omega}_{3}l_{dis} \label{eqloss2}.
\end{align}

\noindent \textbf{4D Point Cloud as Videos through Optimization.} We create multiple \textbf{co-visible} pseudo-cameras ensuring that their optical axes intersect at a single point for modality transformation. The primary pseudo-camera is then selected based on the motion direction and the plane of platform. The intrinsic parameters of pseudo-cameras are selected to match those of the real world camera. For each sequence, the coordinate transformations between all other pseudo-cameras and the primary pseudo-camera are fixed. The extrinsic parameters of primary pseudo-camera are iteratively optimized based on texture and information to select the best observation perspective. 

Taking the right-handed coordinate system as an example, the x-axis represents the forward direction, while the z-axis points upward. For a co-visible pseudo-camera group in a surround-view setup, ideally, when a sufficient number of cameras are available, the six degrees of freedom that affect information and texture reduce to three: $x$, $z$, and $pitch$. We iteratively update the translation and rotation parameters: (1) the translation distance $t$, which moves along a direction perpendicular to the ground, and (2) the rotation angle $\alpha$, whose axis of rotation is orthogonal to both the platform's driving direction and the ground-perpendicular direction. %
\begin{align}
    t_i & = \arg\min_{t_i} h \left( f_{\boldsymbol{\theta}} \left(\xi( \text{Proj} (t \mid \boldsymbol{\alpha}_{i-1}, t_{i-1}, V_{object})) \right) \right) \label{eq:t}, \\
    \boldsymbol{\alpha}_i & = \arg\max_{\boldsymbol{\alpha}_i \in [a, b]} \text{Count} \left( \text{Proj} (\boldsymbol{\alpha} \mid \boldsymbol{\alpha}_{i-1}, t_i, V_{object}) \right) \label{eq:r},
\end{align}
where $\text{Proj}(\cdot)$ is the projection of $V_{object}$ to the pseudo-camera image plane, and $h(\cdot)$ is the distance between the features and the cluster center. $\boldsymbol{\alpha} \in [a, b]$ means optimizing within the visible range of the ground. $\text{Count}(\cdot)$ calculates the number of voxels within the projected image plane.

When solving the discrete optimization problem, we divide the \textit{Keyframes} of the sequence into batches and adopt a greedy optimization strategy as illustrated in Algorithm. \ref{alg:algorithm1}. For each pseudo-image within a batch, we first search the optimal value $t_i$ for each pseudo-image over the entire image plane and then compute its average value to update. Next, we identify the best $ \boldsymbol{\alpha}_i$ for each pseudo-image and average it for the update. Through the above iterative method, our method self-adaptively builds a set of pseudo-cameras which can transfer $V_{object}$ into the same domain with $D_{sample}$ from a texture perspective, while preserving the suboptimal point cloud information. Through the optimized pseudo-cameras, we successfully transform the 4D point cloud into videos.

\begin{algorithm}[H]
\caption{Pseudo Code for Data Alignment}
\label{alg:algorithm1}
\begin{algorithmic}
\STATE 
\STATE \textbf{Input:} SA-V dataset, $V_{object}$
\STATE \textbf{Output:} Aligned data
\STATE \textbf{Pretrain Stage:} 
\STATE \hspace{0.5cm}\# For each VFM, only a single training session is required.
\STATE \hspace{0.5cm}Pseudo-Label generation
\STATE \hspace{0.5cm}Train $f_{\theta}(\cdot)$ with Eq.~\eqref{eqloss1}
\STATE \hspace{0.5cm}Train $f_{\theta}(\cdot)$ with Eq.~\eqref{eqloss2}
\STATE \textbf{Transform Stage:}
\STATE \hspace{0.5cm}\# Optimize for each sequence.
\STATE \hspace{0.5cm}Initialize parameter for primary camera
\WHILE{$t$ and $\boldsymbol{\alpha}$ not converged}
    \STATE \hspace{0.5cm}Do pseudo-color step
    \STATE \hspace{0.5cm}Perform DFT
    \FORALL{$V_{object}$ in batch}
        \STATE \hspace{1.0cm}Calculate Eq.~\eqref{eq:t}
    \ENDFOR
    \STATE \hspace{0.5cm}Update $t$
    \FORALL{$V_{object}$ in batch}
        \STATE \hspace{1.0cm}Calculate Eq.~\eqref{eq:r}
    \ENDFOR
    \STATE \hspace{0.5cm}Update $\boldsymbol{\alpha}$
\ENDWHILE
\STATE Apply projection with $t$ and $\boldsymbol{\alpha}$
\STATE Process data from other pseudo-cameras
\end{algorithmic}
\end{algorithm}

\subsubsection{Pseudo-Color Generation}
\label{colorsec}
In this section, we focus on color generation. For RGB images in $D_{sample}$, natural lighting enhances edge information and when the lighting conditions are similar, the colors of temporal adjacent observations are also similar. For $V_{object}$, we have information with two key properties: \textbf{\textit{Property 1}}: The normalized intensity values of the same material are generally consistent and remain stable across frames. \textbf{\textit{Property 2}}: Depth differences within neighborhood can describe edges. We formulated pseudo-color in HSI format:

\begin{equation}
\begin{split}
        [H,S,I] = [histeq(norm(intensity)), s, \\ {\beta}_{1} + {\beta}_{2} histeq(norm(filter(depth)))],
\end{split}
    \label{eq:hsi}
\end{equation}
where $norm(\cdot)$ is minmax normalization, $histeq(\cdot)$ does histogram equalization and $filter(\cdot)$ calculates differences within neighborhood. Through Eq. \eqref{eq:hsi}, we provide SAM-sensitive edge information to the greatest extent and provide spatiotemporal consistent material color information between frames to facilitate tracking. Finally, we convert the HSI form into the RGB form.

\subsection{Zero-Shot Segmentation with 4D Consistency}
\label{sec3.4}
Compared to SAM~\cite{kirillov2023segment}, SAM2~\cite{ravi2024sam} requires a label to be specified when prompted. We adopt a frame-by-frame automatic prompting approach. Specifically, we employ DBSCAN~\cite{ester1996density} to derive bi-level cluster centers for prompt calculation at \textit{Keyframes}. We match the low-density clustering centers with the high-density clustering centers, and use the high-density clustering centers as the positive prompt. For each positive prompt, its negative prompt comes from the neighbor of the low-density center it matches. The prompt points are then transformed across \textit{Superframes} within the \textit{Keyframe’s} neighborhood through coordinate transformations provided in Sec.~\ref{sec3.2} and are finally projected onto a pseudo-image as 4D-consistent prompt points. Leveraging SAM2’s refinement prompts, the memory bank mechanism, and the \textit{Keyframe’s} informative and representative nature, this method enables us to achieve spatiotemporal consistent prompts across frames. 

We restore the presegmentation results of $Y_{object}$ through geometric transformation. Due to the redundancy of prompt points and the suboptimal tracking performance of SAM2, a merging operation is necessary. Therefore, we introduce a 4D NMS strategy based on traditional 3D NMS. We introduce the Temporal Equivalence Ratio $\Psi$, retaining only the frames where the $\Psi(id1,id2)$ meets the predefined threshold.

\begin{align}
\begin{split}
    \Psi(id1,id2)  &= \tfrac{\sum_f^{F1 \cup F2}EQ(Mask_{id1}^{f},Mask_{id2}^{f})}{min(F1_{max},F2_{max}) - max(F1_{min},F2_{min})},
\end{split}
\label{eq:ter}
\end{align}
where $EQ(\cdot,\cdot)$ means $Mask_{id1}^{f}$ and $Mask_{id2}^{f}$ satisfy the merging condition in 3D NMS at frame $f$. After performing 4D NMS on the \textit{Superframes}, we perform inter-frame smoothing on each single frame. Inter-frame smoothing automatically merges labels whose center point distance and bounding box side length meet strict thresholds.

For $Y_{ground}$, we first project $L_{ground}$ onto a 2D grid map based on the pose. In urban, indoor, and industrial environments, the ground material, which can be reflected by the normalized LiDAR intensity, often conveys underlying semantic information. However, a single normalized intensity value alone is insufficient for distinguishing different surfaces. For instance, dirt roads typically exhibit a high degree of irregularity. Therefore, we use the normalized intensity distribution in its neighborhood as the feature of each grid, and use fuzzy $c$-means~\cite{fcm} for clustering. This approach may lead to over-segmentation, such as distinguishing lane markings from regular road surfaces. However, we consider this outcome beneficial, as merging these segments does not impose a significant additional burden on annotators.

For datasets with calibrated cameras, we use SAM for segmentation and associate point clouds with pixels via extrinsic parameters. Associated points in $L_{ground}$ are grouped by map location into subsets like super pixel. We perform fuzzy $c$-means on these subsets, and the results are used as the initial values of the camera-free mode introduced above. For associated points in $L_{object}$, we split the unsegmented part.

Based on our zero-shot 4D panoptic LiDAR segmentation framework, we build SALT. We leave the details of design of SALT's user interface and user manual in Appendix A. Several implementation details and software acceleration designs are introduced in Appendix B.

\section{Experiments}
\label{sec:exp}

\subsection{Experimental Setup}
\subsubsection{Datasets and Platform}
We evaluate the presegmentation performance of our tool on two \textbf{autonomous driving} benchmarks with mechanical spinning LiDAR: SemanticKITTI~\cite{behley2019semantickitti,behley2021benchmark} (Velodyne HDL-64E with 2 cameras) and nuScenes~\cite{caesar2020nuscenes,fong2022panoptic} (Velodyne HDL-32E with 6 cameras). Moreover, we conducted additional tests on the low-resolution benchmark: SemanticKITTI-16~\cite{yan2024benchmarking} (Reduce LiDAR beams from 64 to 16). Additionally, we assess performance of SALT on the \textbf{industrial robotics} benchmark with hybrid solid LiDAR: S.MID~\cite{wang2024sfpnet} (only Livox Mid-360 data). 

\begin{figure}[t]
    \centering
    \includegraphics[width=1 \linewidth]{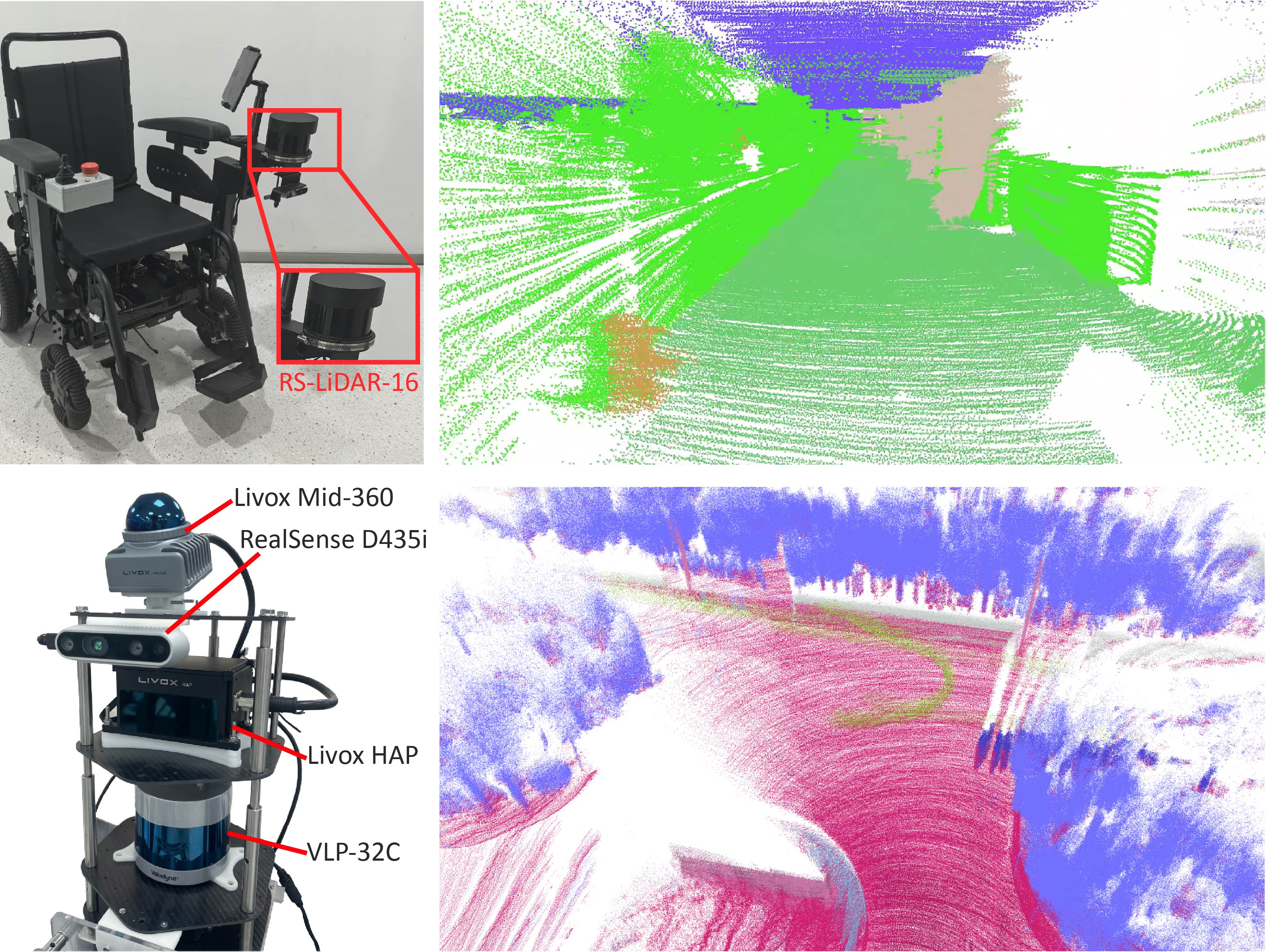}
    \caption{Platform, sensors and cumulated point clouds. Left top: smart wheelchair with 16-beam LiDAR, right top: cumulated indoor point cloud, left bottom: sensor box for combined LiDAR, right bottom: cumulated outdoor point cloud. }
    \label{fig:platform}
\end{figure}

To further validate the zero-shot capabilities and generalizability of our method, we conduct tests on two custom-built platforms: one for annotating \textbf{low-resolution} LiDAR (RS-LiDAR-16) in \textbf{indoor} environments, and another for processing data from \textbf{three combined} LiDAR types (VLP-32C, Livox HAP and Livox Mid-360) in outdoor scenes as shown in Fig.~\ref{fig:platform}. Existing datasets primarily focus on outdoor scenes. To bridge this gap, we employ a smart wheelchair equipped with a mechanical spinning LiDAR to collect raw indoor LiDAR data for additional validation. The indoor sequence comprises 809 frames, annotated into nine classes (\textit{ground, ceiling, wall, column, table, sofa, chair, manmade, human}). Furthermore, existing datasets rarely include multiple LiDAR types. To address this limitation, we utilize an autonomous vehicle equipped with a mechanical spinning LiDAR, a solid-state LiDAR, and a hybrid-solid LiDAR. The collected dataset consists of 1,748 frames, annotated into ten classes (\textit{bush, tree, road, wall, parked bicycles, flower bed, manmade, human, car, cyclist}).

\subsubsection{Metrics and Evaluation}
\label{metric}

\begin{table*}[h]
\caption{Panoptic segmentation on SemanticKITTI and nuScenes validation sets. We compared SALT with fully supervised methods and self-supervised method. Note that all results are obtained from the literature.\label{tab:1}}
\centering
\begin{tabular}{|l|l|l|c|c|c|c|c|c|}
\hline
Dataset & Method & Supervision & $PQ$ & $RQ$ & $SQ$ & $PQ^{St}$ & $PQ^{Th}$ & $mIoU$ \\
\hline
\multirow{9}{*}{SemanticKITTI} 
    & DS-Net ('21) \cite{hong2021lidar} & Full & 57.7 & 68.0 & 77.6 & 61.8 & 54.8 & 63.5 \\
    & PolarSeg ('21) \cite{zhou2021panoptic} & Full & 59.1 & 70.2 & 78.3 & 65.7 & 54.3 & 64.5 \\
    & EfficientLPS ('21) \cite{sirohi2021efficientlps} & Full & 59.2 & 69.8 & 75.0 & 58.0 & 60.9 & 64.9 \\
    & GP-S3Net ('21) \cite{razani2021gp} & Full & 63.3 & 75.9 & 81.4 & 70.2 & 58.3 & 73.0 \\
    & MaskPLS ('23) \cite{marcuzzi2023mask} & Full & 59.8 & 69.0 & 76.3 & - & - & - \\
    & SAL ('24) \cite{ovsep2025better} & Full & 59.5 & 69.2 & 75.7 & 62.3 & 57.4 & 63.8 \\
    & SAL ('24) \cite{ovsep2025better} & Self & 24.8 & 32.3 & 66.8 & 17.4 & 30.2 & 28.7 \\
    & \textbf{Ours} & Zero-shot & 42.8 & 53.5 & 74.1 & 23.5 & 69.4 & 48.5 \\
    & \textbf{Ours with cameras} & Zero-shot & 43.2 & 54.4 & 73.9 & 24.8 & 68.4 & 49.7 \\
\hline
\multirow{9}{*}{nuScenes} 
    & DS-Net ('21) \cite{hong2021lidar} & Full & 51.2 & 59.0 & 86.1 & 38.4 & 72.3 & 73.5 \\
    & GP-S3Net ('21) \cite{razani2021gp} & Full & 61.0 & 72.0 & 84.1 & 56.0 & 66.0 & 75.8 \\
    & PolarSeg ('21) \cite{zhou2021panoptic} & Full & 63.4 & 75.3 & 83.9 & 59.2 & 70.4 & 66.9 \\
    & PHNet ('22) \cite{li2022panoptic} & Full & 74.7 & 84.2 & 88.2 & 74.0 & 75.9 & 79.7 \\
    & MaskPLS ('23) \cite{marcuzzi2023mask} & Full & 57.7 & 66.0 & 71.8 & 64.4 & 52.5 & 62.5 \\
    & SAL ('24) \cite{ovsep2025better} & Full & 70.5 & 80.8 & 85.9 & 79.4 & 61.7 & 72.8 \\
    & SAL ('24) \cite{ovsep2025better} & Self & 38.4 & 47.8 & 77.2 & 47.5 & 29.2 & 33.9 \\
    & \textbf{Ours} & Zero-shot & 38.7 & 48.3 & 79.1 & 40.7 & 36.7 & 28.3 \\
    & \textbf{Ours with cameras} & Zero-shot & 41.4 & 53.7 & 76.4 & 51.6 & 31.2 & 34.1 \\
\hline
\end{tabular}
\end{table*}

Evaluating presegmentation quality essentially involves assessing the quality of panoptic segmentation after alignment with ground-truth labels. We employ standard Panoptic Quality ($PQ$), Segmentation Quality ($SQ$) and mean Intersection over Union ($mIoU$) metrics~\cite{kirillov2019panoptic,ovsep2025better,behley2021benchmark}. Following SAL’s zero-shot evaluation strategy~\cite{ovsep2025better}, we incorporate a semantic oracle (i.e., predicted masks are assigned to ground-truth semantic classes via majority voting) and a stuff-merging approach (i.e., allowing stuff to be split into separate instances). Through the above operations, we assign a unique semantic label to the entire sequence for each pre-segmented index. Then, within each semantic category, we reorder the instance labels based on the pre-segmented index. This assignment process aligns with annotation logic of our tool, making the metric $PQ$ a direct reflection of the annotation workload. Furthermore, to demonstrate the whole sequence consistent tracking capability of our methods, we also employ LiDAR Segmentation and Tracking Quality ($LSTQ$) metrics~\cite{aygun20214d}.

\begin{figure*}[t]
    \centering
    \includegraphics[width=1.0\linewidth]{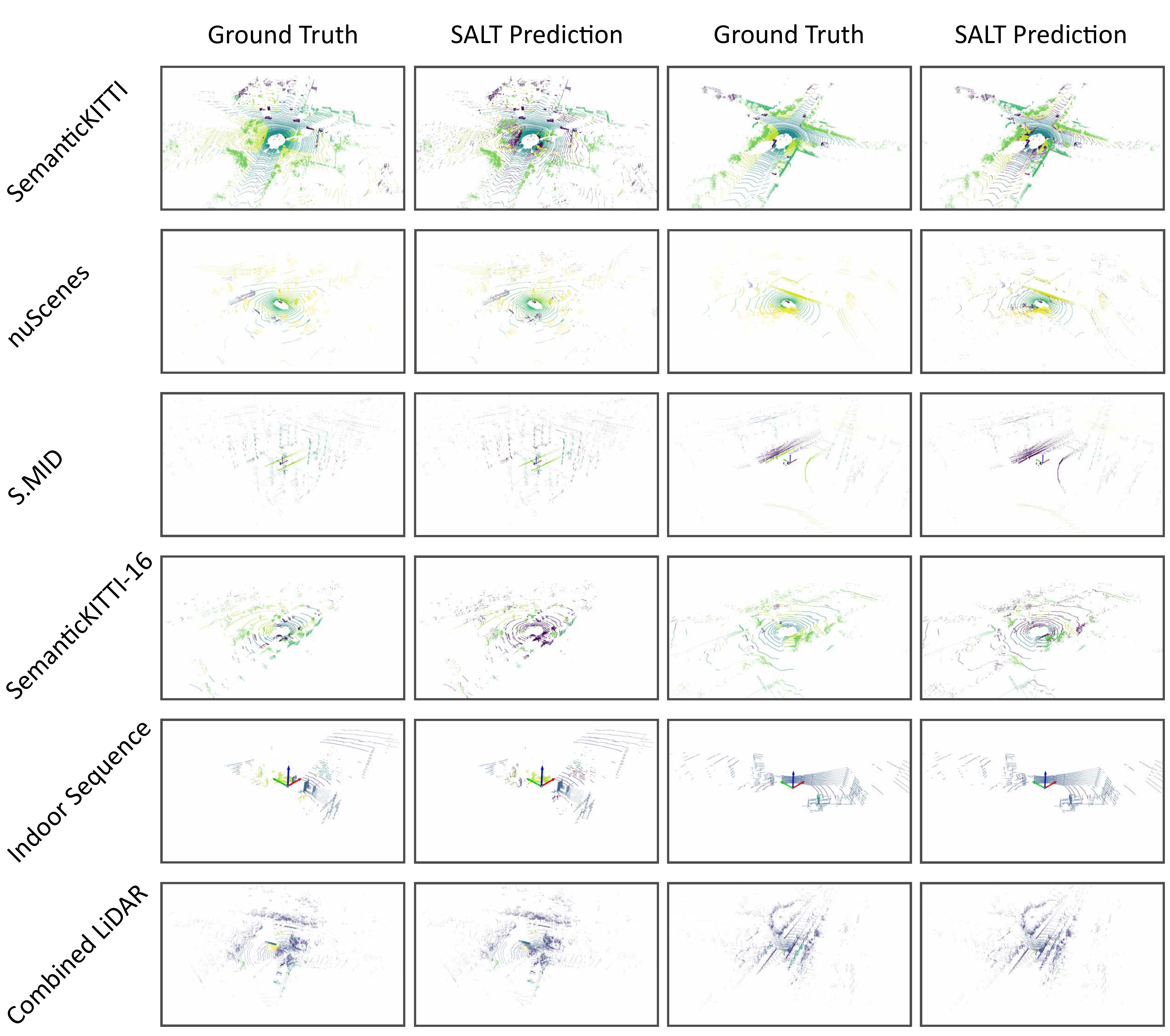}
    \caption{Visualization of LiDAR-only class-agnostic segmentation results with semantic oracle in six datasets with various scenes and different LiDAR setup.}
    \label{fig6:vis}
\end{figure*}

\subsection{Labeling Results}
\textbf{SemanticKITTI.} Segmentation results for SemanticKITTI validation sets are shown in Tab. \ref{tab:1}. Compared to the self-supervised zero-shot method SAL \cite{ovsep2025better}, our approach not only \textbf{eliminates the need for training on this dataset’s LiDAR data} but also achieves a significant \textbf{18.0\%} $PQ$ and \textbf{18.4\%} $mIoU$ improvement. Even compared to supervised methods, our approach achieves 67.6 \~{} 74.2\% of the performance of SOTA methods in terms of $PQ$ in LiDAR-only mode. With the inclusion of the camera modality, our annotation performance improves slightly by 0.4\% in $PQ$ and 1.2\% in $mIoU$. This further confirms that our data alignment approach offers a significant advantage over feature alignment methods in reducing dependence on calibrated cameras, particularly in datasets like SemanticKITTI, where camera data is limited. In particular, if the data collection location lacks lighting conditions (such as at night), existing methods relying on camera distillation will become completely unusable.

4D-consistent Segmentation and Tracking are crucial for annotation tools. As shown in Tab.~\ref{tab:2}, our method achieves 31.5\% $LSTQ$, approximately half the performance of supervised approaches, with significant room for improvement in $S_{assoc}$. Further analysis indicates that the primary limitation stems from the current tracking abilities of SAM2. We anticipate that advancements in VFM will enhance our framework, leading to more effective annotation tools in the future.

\textbf{nuScenes.} In contrast to SemanticKITTI, nuScenes contains more calibrated cameras but more sparse LiDAR data. More camera allows the feature alignment method to perform better. However, our method still outperforms by 0.3\% in $PQ$ with only LiDAR data and 3.0\% with additional camera as shown in Tab.~\ref{tab:1}. Combined with the previous results in SemanticKITTI, it can be seen that as a method developed for LiDAR annotation tools, the performance of our method is more related to the quality of LiDAR data itself.

\begin{table}[h]
\caption{4D Panoptic Segmentation on SemanticKITTI validation set. We compared SALT with fully supervised methods. Note that all results are obtained from the literature.\label{tab:2}}
\centering
\resizebox{1.0\linewidth}{!}{%
\begin{tabular}{|l|l|c|c|c|}
\hline
\textbf{Methods} & Supervision & $LSTQ$ & $S_{assoc}$ & $S_{cls}$ \\
\hline
4D-PLS ('21) \cite{aygun20214d} & Full & 62.7 & 65.1 & 60.5 \\
4D-StOP ('22) \cite{kreuzberg20224d} & Full & 67.0 & 74.4 & 60.3 \\
Eq-4D-StOP ('23) \cite{zhu20234d} & Full & 70.1 & 77.6 & 63.4 \\
Mask4D ('23) \cite{marcuzzi2023mask4d} & Full & 71.4 & 75.4 & 67.5 \\
4D-DS-Net ('24) \cite{hong2024unified} & Full & 68.3 & 71.5 & 65.1 \\
Mask4Former ('24) \cite{yilmaz2024mask4former} & Full & 70.5 & 74.3 & 66.9 \\
\hline
\textbf{Ours} & Zero-shot & 31.5 & 21.5 & 46.1 \\
\hline
\end{tabular}
}
\end{table}

\textbf{S.MID.} S.MID only provides hybrid-solid LiDAR data and corresponding  semantic ground truth labels. Therefore, we only compare the semantic segmentation results in terms of $mIoU$ as shown in Tab.~\ref{tab:smid}. Compared to SemanticKITTI and nuScenes, S.MID presents two key challenges. First, in industrial scenarios, objects with different semantics appear at varying heights in the same location, leading to occlusion and clustering issues. Second, the point cloud distribution of hybrid-solid LiDAR is inherently random, posing significant difficulties even for supervised methods. In such a challenging industrial scenario, our method achieves 64.1 \~{} 68.2\% of the performance of supervised methods, which demonstrate the robustness of our approach to different scenarios and LiDAR types.

\begin{table}[t]
\caption{Semantic Segmentation on S.MID validation set. We compared SALT with fully supervised methods. Note that all results are obtained from the literature.\label{tab:smid}}
\centering
\resizebox{1.0\linewidth}{!}{%
\begin{tabular}{|l|l|c|}
\hline
\textbf{Methods} & Supervision & $mIoU$ \\
\hline
SSCN ('18) \cite{graham20183d} & Full & 67.6 \\
Cylinder3D ('21) \cite{zhu2021cylindrical} & Full & 68.8 \\
SphereFormer ('23) \cite{lai2023spherical} & Full & 67.8 \\
SFPNet ('24) \cite{wang2024sfpnet} & Full & 71.9 \\
\hline
\textbf{Ours} & Zero-shot & 46.1 \\
\hline
\end{tabular}
}
\end{table}

\begin{table}[t]
\caption{Semantic Segmentation on SemanticKITTI-16 validation set. We compared SALT with fully supervised methods. Note that all results are obtained from the literature.\label{tab:kitti16}}
\centering
\resizebox{1.0\linewidth}{!}{%
\begin{tabular}{|l|l|c|}
\hline
\textbf{Methods} & Supervision & $mIoU$ \\
\hline
KPConv ('19) \cite{thomas2019kpconv} & Full & 43.8 \\
MinkowskiNet ('19) \cite{choy20194d} & Full & 50.2 \\
SalsaNext ('20) \cite{cortinhal2020salsanext} & Full & 32.3 \\
\hline
\textbf{Ours} & Zero-shot & 28.2 \\
\hline
\end{tabular}
}
\end{table}

\begin{table}[t]
\caption{Panoptic Segmentation on indoor Low-Resolution LiDAR and outdoor combined LiDAR sets.\label{tab:plat}}
\centering
\resizebox{1.0\linewidth}{!}{%
\begin{tabular}{|c|c|c|c|c|}
\hline
LiDAR & $PQ$ & $RQ$ & $SQ$ & $mIoU$ \\
\hline
Low-Resolution LiDAR (indoor) & 38.0 & 46.3 & 71.4 & 39.2 \\
Combined LiDAR & 52.5 & 66.0 & 76.7 & 54.5 \\
Combined LiDAR with cameras & 54.1 & 69.0 & 76.0 & 57.6 \\
\hline
\end{tabular}
}
\end{table}

\begin{table*}[t]
\caption{A comparison of representative LiDAR point cloud annotation systems. The symbol ``-'' indicates that the evaluation is not applicable. The number of ``+'' represents the performance. ``Scope'' means the scope of automatic annotation by the tool during the user's first interaction with the object or scene. ``ZSL Ability'' means zero-shot ability.\label{tab:labeler}}
\centering
\resizebox{1.0\linewidth}{!}{%
\begin{tabular}{|c|c|c|c|c|c|c|c|}
\hline
& \multicolumn{5}{c|}{Presegmentation} & & \\
\hline
Methods & Automatic & Scope (One click) & Performance & ZSL Ability & Tracking Ability & Visualization & Operation \\
\hline
LABELER \cite{behley2019semantickitti} ('19) & - & - & - & - & - & ++ & ++ \\
SUSTech \cite{sustech} ('20) & $\times$ & Single Object & + & + & + & +++ & +++ \\
Interative4D \cite{interactive4d} ('25) & $\times$ & Single Object & +++ & ++ & +++ & + & + \\
\hline
SALT & $\checkmark$ & Whole Sequence & +++ & +++ & ++ & ++ & ++ \\
\hline
\end{tabular}
}
\end{table*}

\begin{table*}[tb]
  \caption{Ablation study for core optimization module in data alignment.\label{tab:abl}}
  \centering
  \resizebox{1.0\linewidth}{!}{
  \begin{tabular}{|l|l|c|c|c|c|c|c|c|}
    \hline
    & View & $PQ$ & $RQ$ & $SQ$ & $mIoU$ & $LSTQ$ & $S_{assoc}$ & $S_{clc}$ \\
    \hline
    SALT & Pseudo-Camera & 42.8 & 53.5 & 74.1 & 48.5 & 31.5 & 21.5 & 46.1 \\
    \hline
    Ablation 1 & BEV & 20.1 (-22.7) & 29.9 & 66.3 & 29.4 (-19.1) & 5.2 (-26.3) & 1.0 & 28.0 \\
    \hline
    Ablation 2 & Origin View & 19.7 (-23.1) & 27.7 & 68.8 & 34.0 (-14.5) & 14.9 (-16.6) & 7.0 & 32.3 \\
    \hline
  \end{tabular}
  }

\end{table*}

\begin{table}[t]
  \caption{Ablation study for automatic label merging methods.  \label{tab:4dnms}}
  \centering
  \resizebox{1.0\linewidth}{!}{
  \begin{tabular}{|l|l|c|c|c|}
    \hline
     & Merging Operation & $LSTQ$ & $S_{assoc}$ & $S_{clc}$ \\
    \hline
    SALT & 4D NMS + Smoothing & 31.5 & 21.5 & 46.1 \\
    \hline
    Ablation 3 & 3D NMS & 24.4 (-7.1) & 12.7 & 46.9 \\
    \hline
    Ablation 4 & 4D NMS & 27.2 (-4.3) & 15.9 & 46.4 \\
    \hline
    Ablation 5 & 3D NMS + Smoothing & 30.1 (-1.4) & 19.5 & 46.7 \\
    \hline
  \end{tabular}
  }

\end{table}

\textbf{Low-Resolution LiDAR.}
The existing datasets are equipped with LiDAR that has at least 32 channels. However, in practical applications, many devices are equipped with the low-cost, low-resolution LiDAR. Therefore, we also conduct experiments on SemanticKITTI-16. As shown in Tab.~\ref{tab:kitti16}, our method achieves 28.2\% $mIoU$. Although the performance of this challenging experiment decreases compared to the original SemanticKITTI, supervised methods also exhibit degraded performance on the same benchmark. Therefore, SALT still demonstrates a certain level of competitiveness on low-resolution data. Additionally, we collected 16-beam LiDAR data in an indoor environment and compared the presegmentation results with the manually annotated results. As shown in Tab.~\ref{tab:plat}, under such challenging data conditions, our method achieves 38.0\% $PQ$. After validating our method in autonomous driving and industrial scenarios, we further demonstrated its cross-scenario annotation capability in indoor environments.

\textbf{Combined LiDAR sensors.} Many platforms works in safety-critical scenarios are equipped with multiple types of LiDAR simultaneously. The varying characteristics of each LiDAR introduce anisotropic density variations in the point cloud data, posing additional challenges for adaptive annotation tools. The experimental results are shown in Tab.~\ref{tab:plat}, where our method achieves 52.5\% $PQ$ and 54.1\% $PQ$ with camera. SALT achieves nearly 40\~{}50\% of the manual annotation quality across multiple scenarios and various LiDAR types, demonstrating its generalizability.

\textbf{Visualization.}
We visualize the aligned presegmentation results from five datasets in Fig.~\ref{fig6:vis}. We can observe that before the manual annotation, SALT had already completed a large amount of pre-annotation work that was close to the ground truth. Therefore, it can reduce the manual workload and lowered the cost.

\subsection{Ablation Studies}
\label{abl}
In order to evaluate the performance of each design within our tools, we carry out two groups of ablation experiments utilizing the SemanticKITTI validation set as shown in Tab.~\ref{tab:abl} and Tab.~\ref{tab:4dnms}.

\subsubsection{Better Views Yield Closer Textures}
We first validate the effectiveness of our self-supervised distance-based iterative optimization from Eq.~\eqref{eq:t} and Eq.~\eqref{eq:r} in determining a suboptimal projection during modality transformation. We choose the BEV perspective, known for its high informativeness and strong separation, and use a perspective similar to real cameras in the dataset as our baseline. Compared to SALT’s sequentially adaptive suboptimal view, the two commonly used projection views show a significant performance drop: -22.7\% and -23.1\% $PQ$ for panoptic segmentation task and -26.3\% and -16.6\% $LSTQ$ for 4D panoptic LiDAR segmentation as shown in Tab.~\ref{tab:abl}. These results prove that the pseudo-image in SALT aligns more closely with the SA-V dataset distribution than other projections, ultimately leading to better segmentation and tracking performance. 

\subsubsection{4D NMS and Smoothing Make a Good Team}
Remove 4D NMS or inter-frame smoothing operation will hurt 1.4\% and 4.3\% $LSTQ$. Remove both of them and only rely on SAM2's tracking result will hurt 7.1\% $LSTQ$. This demonstrate the effectiveness of Eq.~\eqref{eq:ter} and smoothing operation for 4D panoptic LiDAR segmentation.

\begin{figure}[t]
    \centering
    \includegraphics[width=1 \linewidth]{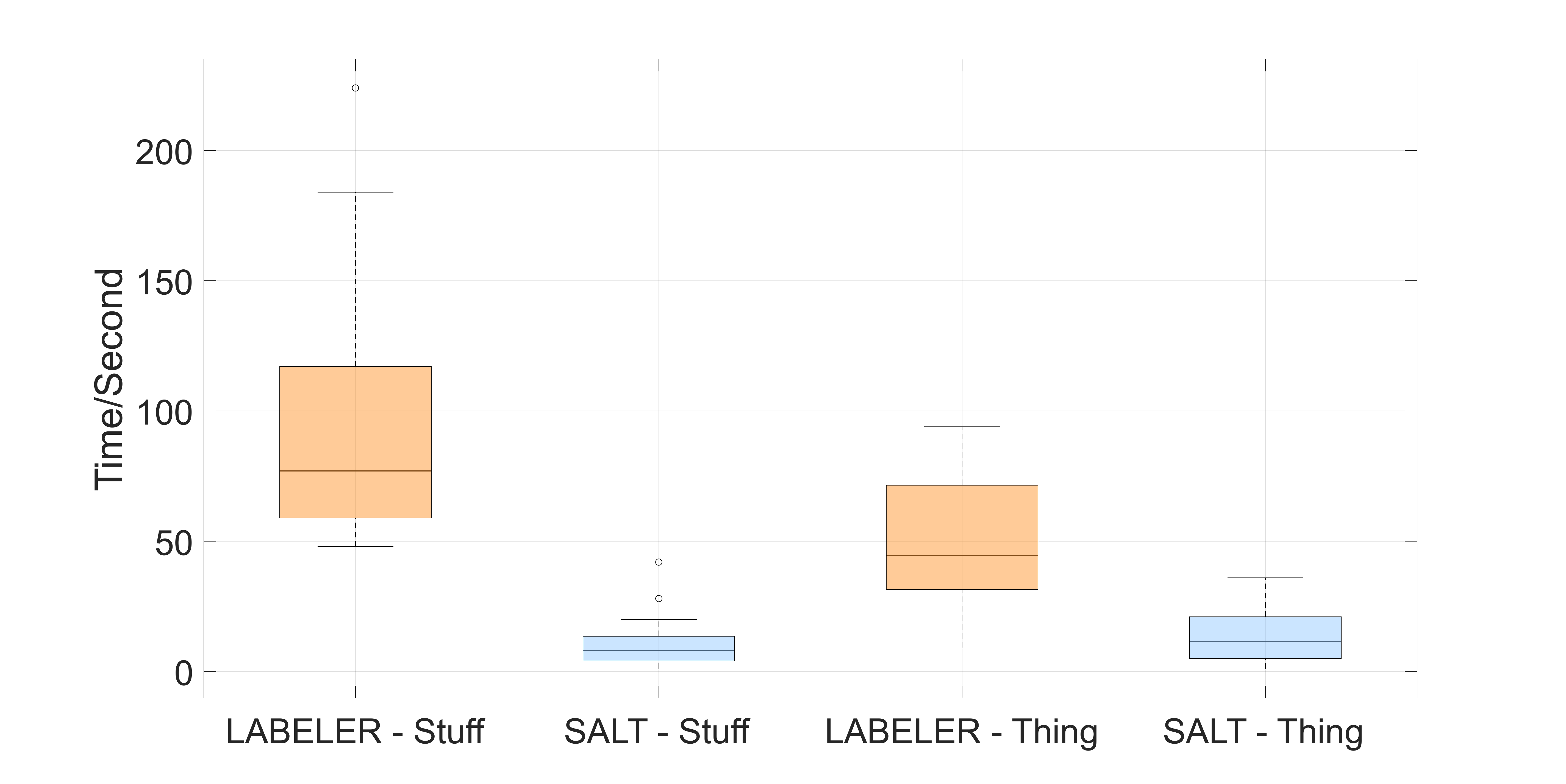}
    \caption{User test results comparison between SALT and LABELER. The annotation time statistics for \textit{stuff} are estimated based on an approximate measurement per unit area of 25 square meters. In contrast, the annotation time statistics for \textit{thing} are calculated based on the number of instances, regardless of their spatial extent.}
    \label{fig:tool}
\end{figure}

\subsection{Tool Discussion}
Our success in zero-shot segmentation has laid a solid foundation for constructing annotation tools. We compared three representative open-source LiDAR annotation tools in Tab.~\ref{tab:labeler}. Our method outperforms existing tools in terms of segmentation performance, tracking quality, visualization, and operation. More importantly, the operation of presegmentation does not require manual interaction, and it can generates the reference for the entire sequence. In addition, the presegmentation methods of other approaches make it difficult to achieve zero-shot segmentation across different types of LiDAR. 

We invited 20 participants to annotate the SemanticKITTI validation set using both SALT and our baseline LABELER. Fig.~\ref{fig:tool} presents a comparison of manual annotation time between SALT and LABELER. In terms of overall annotation efficiency, SALT reduces manual annotation time by approximately 83\%, demonstrating the effectiveness of our tool. Notably, the assignment strategy (the semantic oracle and stuff-merging approach) used for zero-shot segmentation evaluation aligns with the user interaction method (see more details in Appendix A). We found that $PQ$ quality to some extent reflects the degree of manual annotation cost reduction, with noticeable differences between stuff and thing categories. Moreover, the efficiency gap between experienced and inexperienced annotators is significant when using LABELER, whereas SALT is particularly user-friendly for beginners. Additionally, point cloud annotation is often challenging for the human eyes, and regardless of experience level of participants, annotation efficiency typically improves as the task progresses. The presegmentation provided by SALT facilitates the annotation process by offering an initial segmentation, allowing annotators to focus on refinement rather than manual labeling from scratch.

\section{Conclusion}
\label{sec:conc}
We introduce SALT, a flexible semi-automatic labeling tool for general LiDAR point clouds, featuring cross-scene adaptability and 4D consistency. At its core, SALT employs a novel data alignment paradigm that enables seamless modality transformation through refined pseudo-camera projections, bypassing the need for distillation from calibrated real cameras. Our method is further strengthened by a 4D-consistent prompting strategy designed for SAM2 and 4D NMS, ensuring robust segmentation outcomes. SALT demonstrates exceptional zero-shot adaptability across various sensors, scenes, and motion conditions, greatly enhancing annotation efficiency. 

Future improvements can focus on two key aspects. One is refining our data alignment framework by replacing SAM2 with a novel VFM that offers superior tracking performance. Another is addressing the sensitivity of the greedy algorithm to initial values in data alignment. We anticipate that open-sourcing of SALT will scale up LiDAR data accessibility, driving exponential growth in LiDAR datasets and laying a foundation for future LiDAR foundation models. By enabling more accurate and scalable LiDAR perception, these improvements will empower robots to perceive and interact with the physical world in a more robust and intelligent manner.

{\appendices
\section*{Appendix A: User Manual of SALT}
\label{user}

The success of annotation tools~\cite{russell2008labelme} for 2D image has made us aware of the significant contribution that a user-friendly annotation tool can make to the prosperity of the community. We have integrated a fully automatic annotation module into LABELER~\cite{behley2019semantickitti}, as shown in Fig.~\ref{supfig2}, Fig.~\ref{supfig3} and Fig.~\ref{supfig4}. An illustration video can be found in \href{https://drive.google.com/file/d/1Fj8A1lgyYjyhxUSKQeVigt8gU-_ti90i/view?usp=drive_link}{the project video}.

\begin{figure}[t]
    \centering
    \includegraphics[width=0.9 \linewidth]{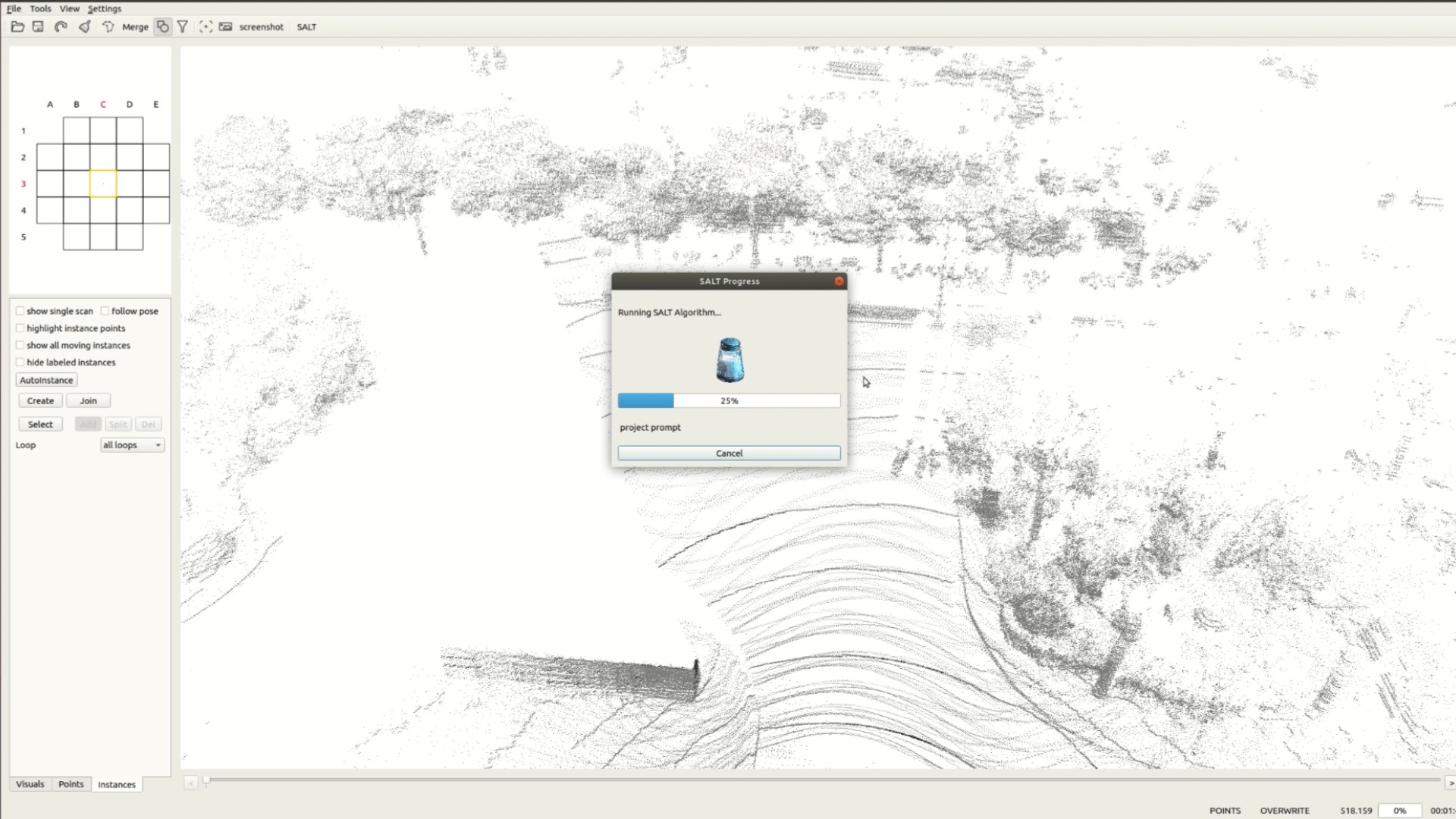}
    \caption{Auto-labeling process in SALT.}
    \label{supfig2}
\end{figure}

\textbf{Automatic Segmentation of Whole Sequence.} After loading the raw point cloud sequence data, the user only needs to click the ``\textit{SALT}" button once to obtain the presegmentation results for the entire sequence as shown in Fig.~\ref{supfig2}. Once the progress bar, which represents the zero-shot segmentation algorithm described in the main text, is complete, the presegmentation results are automatically saved for subsequent semantic and instance labeling. The presegmentation results are also automatically displayed in the user interface with different colors.

\begin{figure}[t]
    \centering
    \includegraphics[width=0.9 \linewidth]{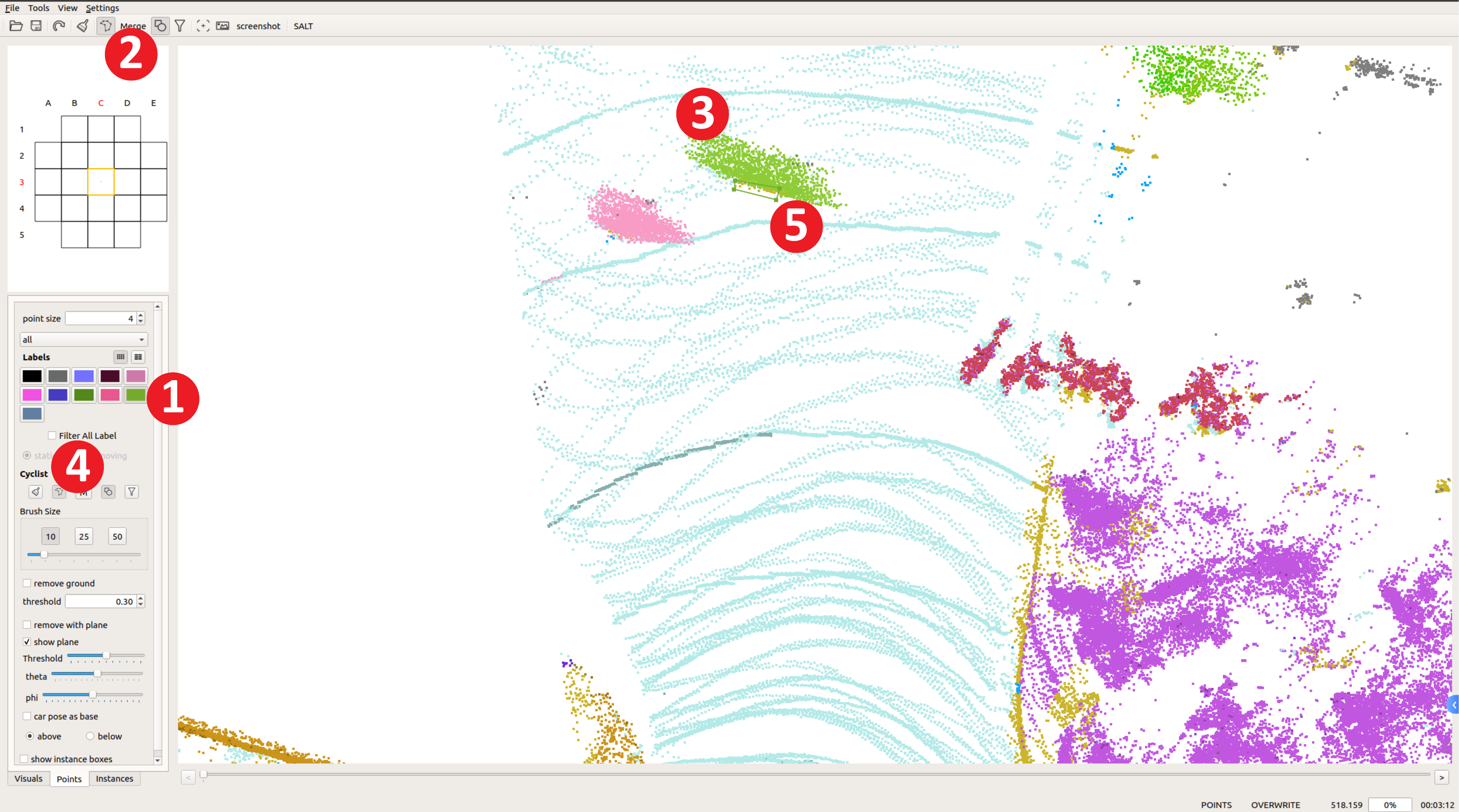}
    \caption{Manual assignment and refinement in SALT for semantic label.}
    \label{supfig3}
\end{figure}

\textbf{Manual Assignment and Refinement for Semantic Annotation.} The user is free to define as many semantic classes appear in the
sequence. Users can assign custom semantic labels to the presegmentation results based on their needs as shown in Fig.~\ref{supfig3}. By simply clicking on a predefined color button representing a specific semantic category and then selecting a point cloud with a particular ID, all points with that ID will be assigned to the chosen label and updated to the corresponding color. This operation is as intuitive and effortless as a coloring game. Please note that the colors used to display the presegmentation results are designed to avoid conflict with user-defined semantic label colors. If users are not satisfied with the pre-annotated results, they can modify them using the polygon tool. Inherited from LABELER, SALT supports the option to hide other classes, making manual annotation adjustments more convenient.

\begin{figure}[t]
    \centering
    \includegraphics[width=0.9 \linewidth]{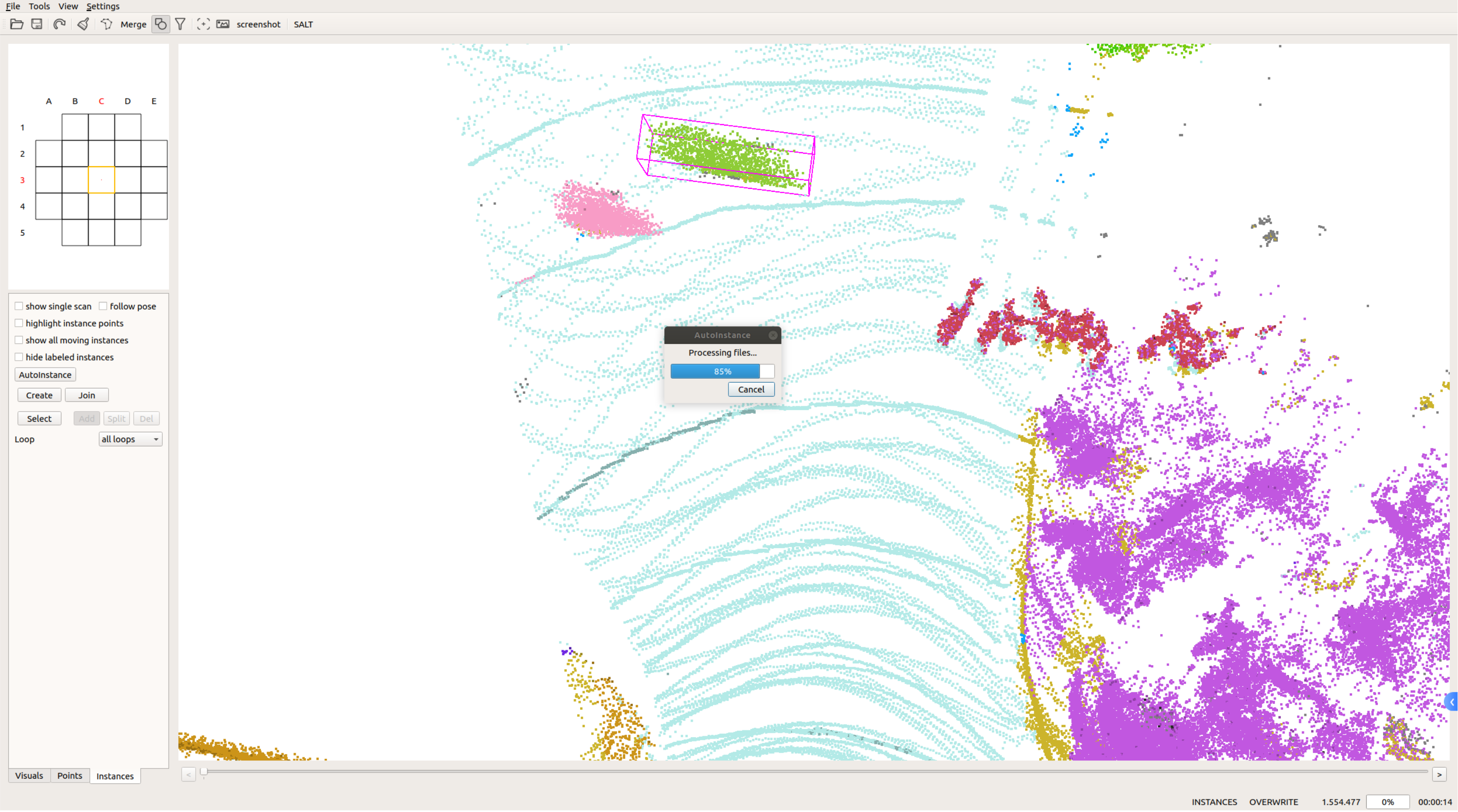}
    \caption{Instance annotation with auto-process and manual refinement in SALT.}
    \label{supfig4}
\end{figure}

\textbf{Auto Ordering and Manual Refinement for Instance Annotation.} Once users are satisfied with the semantic labeling results, they can simply click the ``\textit{Auto Instance}" button to automatically assign instance IDs to all semantic categories as shown in Fig.~\ref{supfig4}. Users can then further refine the results by splitting or merging instance IDs within each category.

\section*{Appendix B: Library of SALT}
\label{code}
\textbf{Projection for Pseudo-Image.} LiDAR Point clouds are inherently sparse, and directly projecting the original point cloud into an pseudo-image typically results in a lack of continuity. A straightforward approach is to voxelize the point cloud and fill the voxels with point data for projection. Assume that each voxel is filled with $m$ points (a parameter related to the pseudo-camera’s intrinsic parameters). In general, $m > 10^6$, and the time complexity for rendering an image in this manner is $O(mv)$, where $v$ represents the number of voxels. Considering the characteristics of LiDAR data, we proposes an accelerated image rendering technique. For each voxel, only the 8 corner points are considered. These corner points are projected onto the image plane, and we construct a convex hull. The pixels within the convex hull are then rendered. The time complexity of this method is $O(8v)$, significantly reducing rendering time.

\textbf{Unprojection for Presegmentation Results.} Point cloud growth is employed to reconstruct the results after segmented and tracked by SAM2, while 3D NMS serves as the foundation for the 4D NMS introduced in the main text. The corresponding algorithm is shown in Algorithm.~\ref{alg:2}.

\begin{algorithm}[t]
\caption{Point Cloud Growth and NMS for Segmentation}
\label{alg:2}
\begin{algorithmic}
\STATE \textbf{Input:} $L_{object}$, Mask
\STATE \textbf{Output:} $Y_{object}^{3dnms}$

\STATE \textit{Step 1: Unprojection based on the given mask.}
\STATE Partial\_segmented\_pointcloud $\gets$ Unproject($L_{object}$, Mask)
\STATE Partial\_segmented\_voxels $\gets$ Mapping(Partial\_segmented \_pointcloud)
\STATE \textit{Step 2: Perform region growth to get segmented voxel clusters.}
\STATE Partial\_segmented\_clusters $\gets$ Region\_growth(Partial \_segmented\_voxels)

\STATE \textit{Step 3: Process each partial segmented cluster individually.}
\FOR{each Cluster in Partial\_segmented\_clusters}
    \STATE \textit{Step 3.1: Refine cluster’s labels using the reduce bleeding operation.}
    \STATE Refined\_voxels $\gets$ Nerf\_bleeding(Cluster)
    
    \STATE \textit{Step 3.2: Perform 3D NMS based on bounding boxes.}
    \STATE Refined\_voxels $\gets$ NMS3d (Refined\_voxels)
    
    \STATE \textit{Step 3.3: Expand labels to unlabeled voxels, prioritizing the most frequent label.}
    \STATE Refined\_voxels $\gets$ Label\_growth (Refined\_voxels)
\ENDFOR

\STATE \textit{Step 4: Convert the refined voxel representation into final labels.}
\STATE $Y_{object}^{3dnms}$ $\gets$ Voxel\_to\_label (Refined\_voxels)

\end{algorithmic}
\end{algorithm}

\textbf{Parallelizing SAM2 Inference for Efficiency.} Given that SAM2 exhibits a relatively low frame rate, we propose a solution by leveraging multi-process parallelism to accelerate its inference. We perform asynchronous inference for each prompt, while also maximizing the usage of the GPU memory to ensure that computational units are fully utilized. This simple approach significantly reduces the processing time.}

\bibliographystyle{IEEEtran}
\bibliography{main}

\vfill

\end{document}